\pdfoutput=1

\documentclass[11pt]{article}

\usepackage{emnlp2021}

\usepackage{times}
\usepackage{latexsym}

\usepackage[T1]{fontenc}

\usepackage[utf8]{inputenc}

\usepackage{microtype}

\usepackage{booktabs}
\usepackage{amsmath}
\usepackage{amssymb}
\usepackage{graphics}
\usepackage{graphicx}
\usepackage{multirow}
\usepackage[ruled,vlined,linesnumbered]{algorithm2e}

\DeclareMathOperator*{\argmin}{argmin}
\DeclareMathOperator*{\argmax}{argmax}

\title{Competence-based Curriculum Learning for \\  Multilingual Machine Translation}

\author{
  Mingliang Zhang\textsuperscript{1}\thanks{ \ \ Work was done when Mingliang Zhang was interning at Pattern Recognition Center, WeChat AI, Tencent Inc, China.}~, 
  Fandong Meng\textsuperscript{2},
  Yunhai Tong\textsuperscript{1}, 
  Jie Zhou\textsuperscript{2}\\
  
  \textsuperscript{1}Key Laboratory of Machine Perception, School of EECS, Peking University \\
  \textsuperscript{2}Pattern Recognition Center, WeChat AI, Tencent Inc, China \\
  \texttt{\{zml24,yhtong\}@pku.edu.cn} \\  \texttt{\{fandongmeng,withtomzhou\}@tencent.com} \\
}

\begin{document}
\maketitle
\begin{abstract}

Currently, multilingual machine translation is receiving more and more attention since it brings better performance for low resource languages (LRLs) and saves more space.
However, existing multilingual machine translation models face a severe challenge: imbalance.
As a result, the translation performance of different languages in multilingual translation models are quite different.
We argue that this imbalance problem stems from the different learning competencies of different languages.
Therefore, we focus on balancing the learning competencies of different languages and propose \emph{\textbf{C}ompetence-based \textbf{C}urriculum \textbf{L}earning for \textbf{M}ultilingual Machine Translation}, named CCL-M.
Specifically, we firstly define two competencies to help schedule the high resource languages (HRLs) and the low resource languages: 1) \emph{Self-evaluated Competence}, evaluating how well the language itself has been learned; and 2) \emph{HRLs-evaluated Competence}, evaluating whether an LRL is ready to be learned according to HRLs’ \emph{Self-evaluated Competence}.
Based on the above competencies, we utilize the proposed CCL-M algorithm to gradually add new languages into the training set in a curriculum learning manner.
Furthermore, we propose a novel competence-aware dynamic balancing sampling strategy for better selecting training samples in multilingual training.
Experimental results show that our approach has achieved a steady and significant performance gain compared to the previous state-of-the-art approach on the TED talks dataset.

\end{abstract}

\section{Introduction} \label{sec:intro}

With the development of natural language processing and deep learning, multilingual machine translation has gradually attracted the interest of researchers \citep{dabre-etal-2020-multilingual}.
Moreover, the multilingual machine translation model demands less space than multiple bilingual unidirectional machine translation models, making it more popular among developers \citep{liu2020multilingual, zhang-etal-2020-improving, fan2020beyond}.

However, existing multilingual machine translation models face imbalance problems.
On the one hand, various sizes of training corpora for different language pairs cause imbalance.
Typically, the training corpora size of some high resource languages (HRLs) is hundreds or thousands of times that of some low resource languages (LRLs) \citep{schwenk2019ccmatrix}, resulting in lower competence of LRL learning.
On the other hand, translation between different languages has different difficulty, which also leads to imbalance. 
In general, translation between closely related language pairs is more effortless than that between distant language pairs, even if the training corpora is of the same size \citep{barrault-etal-2020-findings}.
This would lead to low learning competencies for distant languages compared to closely related languages.
Therefore, multilingual machine translation is inherently imbalanced, and dealing with this imbalance is critical to advancing multilingual machine translation \citep{dabre-etal-2020-multilingual}.

To address the above problem, existing balancing methods can be divided into two categories, i.e., static and dynamic. 
1) Among static balancing methods, temperature-based sampling \citep{arivazhagan2019massively} is the most common one, compensating for the gap between different training corpora sizes by oversampling the LRLs and undersampling the HRLs.
2) Researchers have also proposed some dynamic balancing methods \citep{jean2019adaptive, wang-etal-2020-balancing}.
\citet{jean2019adaptive} introduce an adaptive scheduling, oversampling the languages with poorer results than their respective baselines.
In addition, MultiDDS-S \citep{wang-etal-2020-balancing} focus on learning an optimal strategy to automatically balance the usage of training corpora for different languages at multilingual training.

Nevertheless, the above methods focus too much on balancing LRLs, resulting in lower competencies for HRLs compared to that trained only on bitext corpora.
Consequently, the performance on the HRLs by the multilingual translation model is inevitably worse than that of bitext models by a large margin \citep{lin-etal-2020-pre}.
Besides, knowledge learned by related HRLs is also beneficial for LRLs \citep{neubig-hu-2018-rapid}, while is neglected by previous approaches, limiting the performance on LRLs.

Therefore, in this paper, we try to balance the learning competencies of languages and propose a \emph{\textbf{C}ompetence-based \textbf{C}urriculum \textbf{L}earning Approach for \textbf{M}ultilingual Machine Translation}, named CCL-M. 
Specifically, we firstly define two competence-based evaluation metrics to help schedule languages, which are 1) \emph{Self-evaluated Competence}, for evaluating how well the language itself has been learned; and 2) \emph{HRLs-evaluated Competence}, for evaluating whether an LRL is ready to be learned by the LRL-specific HRLs' \emph{Self-evaluated Competence}. 
Based on the above two competence-based evaluation metrics, we design the CCL-M algorithm to gradually add new languages into the training set.
Furthermore, we propose a novel competence-aware dynamic balancing sampling method for better selecting training samples at multilingual training. 

We evaluate our approach on the multilingual Transformer \citep{vaswani2017attention} and conduct experiments on the TED talks\footnote{\url{https://www.ted.com/participate/translate}} to validate the performance in two multilingual machine translation scenarios, i.e., \emph{many-to-one} and \emph{one-to-many} ("\emph{one}" refers to English).
Experimental results show that our approach brings in consistent and significant improvements compared to the previous state-of-the-art approach \citep{wang-etal-2020-balancing} on multiple translation directions in the two scenarios.

Our contributions\footnote{We release our code on \url{https://github.com/zml24/ccl-m}.} are summarized as follows:
\begin{itemize}
\item 
We propose a novel competence-based curriculum learning method for multilingual machine translation.
To the best of our knowledge, we are the first that integrate curriculum learning into multilingual machine translation.
\item
We propose two effective competence-based evaluation metrics to dynamically schedule which languages to learn, and a competence-aware dynamic balancing sampling method for better selecting training samples at multilingual training.
\item 
Comprehensive experiments on the TED talks dataset in two multilingual machine translation scenarios, i.e., \emph{many-to-one} and \emph{one-to-many}, demonstrating the effectiveness and superiority of our approach,
which significantly outperforms the previous state-of-the-art approach.
\end{itemize}

\section{Background}

\subsection{Multilingual Machine Translation}
Bilingual machine translation model translates a sentence of source language $S$ into a sentence of target language $T$ (\citealp{sutskever2014sequence}; \citealp{cho-etal-2014-learning}; \citealp{bahdanau2014neural}; \citealp{luong-etal-2015-effective}; \citealp{vaswani2017attention}), which is trained as
\begin{equation}
\theta^* = \argmin_\theta \mathcal{L} (\theta; S, T) ,
\end{equation}
where $\mathcal{L}$ is the loss function, $\theta^*$ is the model parameters.

Multilingual machine translation system aims to train multiple language pairs in a single model, including \emph{many-to-one} (translation from multiple languages into one language), \emph{one-to-many} (translation from one language to multiple languages), and \emph{many-to-many} (translation from several languages into multiple languages) \citep{dabre-etal-2020-multilingual}.
Specifically, we denote the training corpora of $n$ language pairs in multilingual machine translation as $\{S_1, T_1\}$, $\{S_2, T_2\}$, $\dots$, $\{S_n, T_n\}$ and multilingual machine translation aims to train a model $\theta^*$ as
\begin{equation}
\theta^* = \argmin_\theta \frac{1}{n} \sum_{i = 1}^n \mathcal{L} (\theta; S_i, T_i) .
\end{equation}

\subsection{Sampling Methods}

Generally, the size of the training corpora for different language pairs in multilingual machine translation varies greatly.
Researchers hence developed two kinds of sampling methods, i.e., static and dynamic, to sample the language pairs at training \citep{dabre-etal-2020-multilingual}.

There are three mainstream static sampling methods, i.e., uniform sampling, proportional sampling, and temperature-based sampling \citep{arivazhagan2019massively}.
These methods sample the language pairs by the predefined fixed sampling weights $\psi$.

\paragraph{Uniform Sampling.} Uniform sampling is the most straightforward solution \citep{johnson-etal-2017-googles}. The sampling weight $\psi_i$ for each language pair $i$ of this method is calculated as follows
\begin{equation}
\psi_i = \frac{1}{\vert \mathcal{S}_\text{lang} \vert ,}
\end{equation}
where $\mathcal{S}_\text{lang}$ is the language sets for training.

\paragraph{Proportional Sampling.} Another method is sampling by proportion \citep{neubig-hu-2018-rapid}. This method improves the model's performance on high resource languages and reduces the performance of the model on low resource languages. Specifically, we calculate its sampling weight $\psi_i$ for each language pair $i$ as 
\begin{equation}
\psi_i = \frac{\vert \mathcal{D}^i_\text{Train} \vert}{\sum_{k \in \mathcal{S}_\text{lang} } \vert \mathcal{D}^k_\text{Train} \vert} ,
\end{equation}
where $\mathcal{D}_\text{Train}$ is the training corpora of language $i$.

\paragraph{Temperature-based Sampling.} It samples the language pairs according to the corpora size exponentiated by a temperature term $\tau$ (\citealp{arivazhagan2019massively}; \citealp{conneau-etal-2020-unsupervised}) as 
\begin{equation}
\psi_i = \frac{p_k^{1 / \tau}}{\sum_{k \in \mathcal{S}_\text{lang}} p_k^{1 / \tau}} \ \text{where} \ p_i = \frac{\vert \mathcal{D}^i_\text{Train} \vert}{\sum_{k \in \mathcal{S}_\text{lang} } \vert \mathcal{D}^k_\text{Train} \vert} .
\end{equation}
Obviously, $\tau = \infty$ is the uniform sampling and $\tau = 1$ is the proportional sampling. Both of them are a bit extreme from the perspective of $\tau$.
In practice, we usually select a proper $\tau$ to achieve a balanced result.

On the contrary, dynamic sampling methods (e.g., MultiDDS-S\citep{wang-etal-2020-balancing}) aim to automatically adjust the sampling weights by some predefined rules.

\paragraph{MultiDDS-S.} MultiDDS-S \citep{wang-etal-2020-balancing} is a dynamic sampling method performing differentiable data sampling.
It takes turns to optimize the sampling weights of different languages and the multilingual machine translation model, showing more significant potential than static sampling methods. This method optimizes the sample weight $\psi$ to minimize the development loss as follows
\begin{equation}
\psi^* = \argmin_\psi \mathcal{L} (\theta^*; \mathcal{D}_\text{Dev}) , \\
\end{equation}
\begin{equation}
\theta^* = \argmin_\theta \sum_{i = 1}^n \psi_i \mathcal{L} (\theta; \mathcal{D}_\text{Train}) ,
\end{equation}
where $\mathcal{D}_\text{Dev}$ and $\mathcal{D}_\text{Train}$ denote the development corpora and the training corpora, respectively.

\section{Methodology}

In this section, we first define a directed bipartite language graph, on which we deploy the languages to train.
Then, we define two competence-based evaluation metrics, i.e., the \emph{Self-evaluated Competence} $c$ and the \emph{HRLs-evaluated Competence} $\hat{c}$, to help decide which languages to learn.
Finally, we elaborate the entire CCL-M algorithm.

\subsection{Directed Bipartite Language Graph}

Formally, we define a directed bipartite language graph $G(V, E)$, in which one side is full of HRLs and the other side of LRLs.
Each vertex $v_i$ on the graph represents a language, and the weight of each directed edge (from HRLs to LRLs) $e_{ij}$ indicates the similarity between a HRL $i$ and an LRL $j$: 
\begin{equation}
e_{ij} = \text{sim}(i, j) .
\end{equation}
Inspired by TCS \citep{wang-neubig-2019-target}, we measure it using vocabulary overlap and define the language similarity between language $i$ and language $j$ as
\begin{equation}
\text{sim}(i, j) = \frac{\vert \text{vocab}_k(i) \cap \text{vocab}_k(j) \vert}{k} ,
\label{eq:sim}
\end{equation}
where $\text{vocab}_k(\cdot)$ represents the top $k$ most frequent subwords in the training corpus of a specific language.

\begin{figure*}[!t]
\centering
\includegraphics[width=0.9\textwidth]{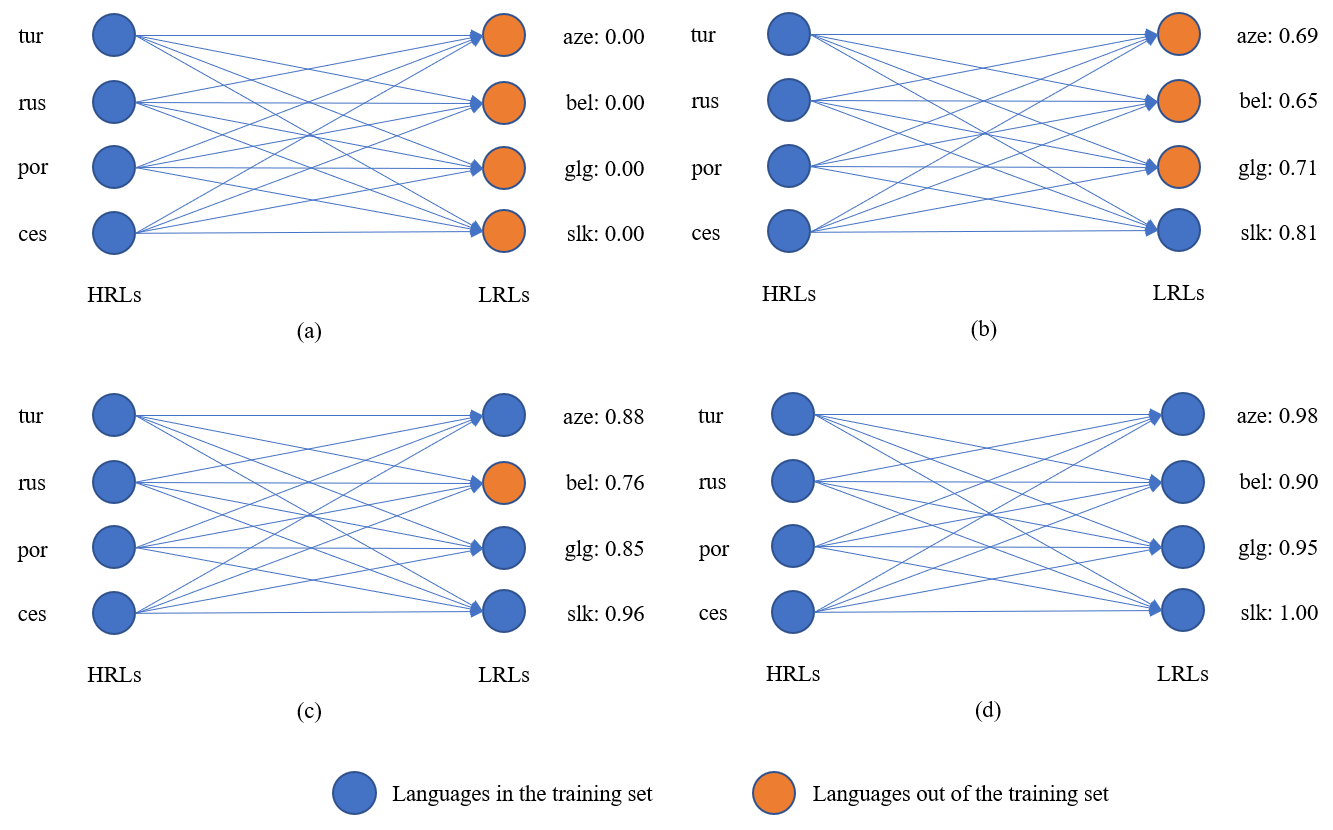}
\caption{Diagram of the CCL-M Algorithm. This graph shows how to gradually add the LRLs to the training set $\mathcal{S}_\text{selected}$ using graph coloring. "aze" stands for Azerbaijani, "bel" stands for Belarusian, etc. The number after the colon indicates current \emph{HRLs-evaluated Competence}, and suppose the corresponding threshold $t$ is set to 0.8. Subfigure (a) represents the state before training. Subfigure (b) indicates that "slk" (Slovak) is added to the training set because the \emph{HRLs-evaluated Competence} is higher than the threshold. Subfigure (c) indicates that "aze" (Azerbaijani) and "glg" (Glacian) are added to the training set, and Subfigure (d) indicates that all the LRLs are added to the training set. Notice we use the abbreviation of language (xxx) to indicate language pairs (xxx-eng or eng-xxx), which is more general.} 
\label{ccl-m}
\end{figure*}

\subsection{Competence-based Evaluation Metrics}

\paragraph{Self-evaluated Competence.} We define how well a language itself has been learned as the \emph{Self-evaluated Competence} $c$.
In the following paragraphs, we first introduce the concept of \emph{Likelihood Score} and then give a formula for calculating the \emph{Self-evaluated Competence} in multilingual training based on the relationship between current \emph{Likelihood Score} and the \emph{Likelihood Score} of model trained on bitext corpus.

For machine translation, we usually use the label smoothed \citep{szegedy2016rethinking} cross-entropy loss $\mathcal{L}$ to measure how well the model is trained, and calculate it as
\begin{equation}
\mathcal{L} = - \sum_i p_i \log_2 q_i ,
\end{equation}
where $p$ is the label smoothed actual probability distribution, and $q$ is the model output probability distribution\footnote{We select 2 as the base number for all relevant formulas and experiments in this paper.}.

We find that the exponential of negative label smoothed cross-entropy loss is a likelihood to some extent, which is negatively correlated to the loss.
Since neural network usually optimizes the model by minimizing the loss, we use the likelihood as a positive correlation indicator to measure competence.
Therefore, we define a \emph{Likelihood Score} $s$ to estimate how well the model is trained as follows
\begin{equation}
s = 2^{-\mathcal{L}} = \prod_i q_i^{p_i} .
\end{equation}

Inspired by \citet{jean2019adaptive}, we estimate the \emph{Self-evaluated Competence} $c$ of a specific language by calculating the quotient of its current \emph{Likelihood Score} and baseline's \emph{Likelihood Score}.
Finally, we obtain the formula as follows
\begin{equation} \label{self-competence}
c = \frac{s}{s^*} = 2^{\mathcal{L}^* - \mathcal{L}} ,
\end{equation}
where $\mathcal{L}$ is the current loss on the development set, $\mathcal{L}^*$ is the \emph{benchmark}  loss of the converged bitext model on the development set, and $s$ and $s^*$ are their corresponding \emph{Likelihood Scores}, respectively.


\paragraph{HRLs-evaluated Competence.} Furthermore, we define how well an LRL is ready to be learned as its \emph{HRLs-evaluated Competence} $\hat{c}$.
We believe that each LRL can learn adequate knowledge from its similar HRLs before training.
Therefore, we estimate each LRL's \emph{HRLs-evaluated Competence} by the LRL-specific HRLs' \emph{Self-evaluated Competence}.

Specifically, we propose two methods for calculating the \emph{HRLs-evaluated Competence}, i.e., \emph{maximal} ($\text{CCL-M}_\text{max}$) and \emph{weighted average} ($\text{CCL-M}_\text{avg}$).
The $\text{CCL-M}_\text{max}$ only migrates the knowledge from the HRL that is most similar to the LRL, so we calculate \emph{maximal} \emph{HRLs-evaluated Competence} $\hat{c}_{\text{max}}$ for each LRL $j$ as
\begin{equation}
\hat{c}_{\text{max}}(j)= c_{\argmax_{i \in \mathcal{S}_\text{HRLs}} e_{ij}} ,
\end{equation}
where $\mathcal{S}_{\text{HRLs}}$ is the set of the HRLs. 

On the other side, the $\text{CCL-M}_\text{avg}$ method pays attention to all HRLs.
In general, the higher the language similarity, the more knowledge an LRL can migrate from HRLs.
Therefore, we calculate \emph{weighted average} \emph{HRLs-evaluated Competence} $\hat{c}_{\text{avg}}$ for each LRL $j$ as
\begin{equation}
\hat{c}_{\text{avg}}(j)= \sum_{i \in \mathcal{S}_\text{HRLs}} \left ( \frac{e_{ij}}{\sum_{k \in \mathcal{S}_\text{HRLs}} e_{kj}} \cdot c_i \right ) .
\end{equation}

\subsection{The CCL-M Algorithm}

Now we detailly describe the \emph{\textbf{C}ompetence-based \textbf{C}urriculum \textbf{L}earning for \textbf{M}ultilingual Machine Translation}, namely the CCL-M algorithm. The algorithm is divided into two parts: 1) curriculum learning scheduling framework, guiding when to add a language to the training set; 2) competence-aware dynamic balancing sampling, guiding how to sample languages in the training set.

First, we present how to schedule which languages on the directed bipartite language graph should be added to the training set  according to the two competence-based evaluation metrics as shown in Figure \ref{ccl-m} and Algorithm \ref{alg:the_alg}, where $\mathcal{S}_{\text{LRLs}}$ is the set of LRLs, and $f(\cdot)$ is the function calculating the \emph{HRLs-evaluated Competence} $\hat{c}$ for LRLs.
Initialized as Line \ref{lst:1}, we add all languages on the HRLs side to the training set $\mathcal{S}_\text{selected}$ at the beginning of training, leaving all languages on the LRLs side in the candidate set $\mathcal{S}_\text{candidate}$.
Then, we regularly sample the development corpora of different languages and calculate current \emph{HRLs-evaluated Competence} of the languages in the candidate set $\mathcal{S}_\text{candidate}$ as shown in Line \ref{lst:8} and \ref{lst:9}.
Further, the "if" condition in Line \ref{lst:13} illustrates that we would add the LRL to the training set $\mathcal{S}_\text{selected}$ when its \emph{HRLs-evaluated Competence} is greater than a pre-defined threshold $t$.
However, as the calculation of Equation \ref{self-competence}, the upper bound of the \emph{Self-evaluated Competence} for a specific language may not always be 1 at multilingual training.
This may cause that some LRLs remain out of the training set $\mathcal{S}_\text{selected}$ for some thresholds.
To ensure the completeness of our algorithm, we will directly add the languages still in the candidate set $\mathcal{S}_\text{candidate}$ to the training set $\mathcal{S}_\text{selected}$ after a long enough number of steps, which is described between Line \ref{lst:22} and Line \ref{lst:32}.


\begin{algorithm}[!t]
\SetAlgoLined
\KwIn{Randomly initialized model $\theta$; language graph $G$; \emph{benchmark} losses $\mathcal{L}_i^*$; training corpora $\mathcal{D}_\text{Train}$; development corpora $\mathcal{D}_\text{Dev}$;}
\KwOut{The converged model $\theta^*$;}
$\mathcal{S}_\text{selected} \gets \mathcal{S}_\text{HRLs}$, $\mathcal{S}_\text{candidate} \gets \mathcal{S}_\text{LRLs}$,
$\psi \gets 0$\; \label{lst:1}
\For{$i \in \mathcal{S}_\text{\normalfont{selected}}$}{
    $\psi_i \gets \frac{1}{\vert \mathcal{S}_\text{\normalfont{selected}} \vert}$\; \label{lst:3}
}
\While{$\theta$ \normalfont{not converge}}{
    train the model on $\mathcal{D}_\text{Train}$ for some steps with sampling weight $\psi$\;
    \For{$i \in \mathcal{S}_\text{\normalfont{selected}} \cup \mathcal{S}_\text{\normalfont{candidate}}$}{
        sample $\mathcal{D}_\text{Dev}$ and calculate $\mathcal{L}_i$\; \label{lst:8}
        $c_i \gets 2^{\mathcal{L}_i^* - \mathcal{L}_i}$\; \label{lst:9}
    }
    \For{$i \in \mathcal{S}_\text{\normalfont{candidate}}$}{
        $\hat{c}_i \gets f(G, i, c_{\mathcal{S}_\text{HRLs}})$\;
        \If{$\hat{c}_i \geq t$}{ \label{lst:13}
            $\mathcal{S}_\text{selected} \gets \mathcal{S}_\text{selected} \cup \{ i \}$\;
            $\mathcal{S}_\text{candidate} \gets \mathcal{S}_\text{candidate} \setminus \{ i \} $\;
        }
    }
    \For{$i \in \mathcal{S}_\text{\normalfont{selected}}$}{
        $\psi_i \gets \frac{1}{c_i}$\;
    }
 }
\If{$\mathcal{S}_\text{\normalfont{candidate}} \neq \varnothing$}{ \label{lst:22}
    $\mathcal{S}_\text{selected} \gets \mathcal{S}_\text{selected} \cup \mathcal{S}_\text{candidate}$\;
    \While{$\theta$ \normalfont{not converge}}{
        train the model on $\mathcal{D}_\text{Train}$ for some steps with sampling weight $\psi$\;
        \For{$i \in \mathcal{S}_\text{\normalfont{selected}}$}{
            sample $\mathcal{D}_\text{Dev}$ and calculate $\mathcal{L}_i$\;
            $c_i \gets 2^{\mathcal{L}_i^* - \mathcal{L}_i}$\;
            $\psi_i \gets \frac{1}{c_i}$\;
        }
    }
} \label{lst:32}
\caption{The CCL-M Algorithm}
\label{alg:the_alg}
\end{algorithm}

Then, we introduce our competence-aware dynamic balancing sampling method, which is based on the \emph{Self-evaluated Competence}.
For languages in the training set $\mathcal{S}_\text{selected}$, we randomly select samples from the development corpora and calculate their \emph{Self-evaluated Competence}.
Those languages with low \emph{Self-evaluated Competence} should get more attention, therefore we simply assign the sampling weight $\psi_i$ to each language $i$ in the training set to the reciprocal of its \emph{Self-evaluated Competence}, as follows 
\begin{equation}
\psi_i \propto \frac{1}{c_i} = 2^{\mathcal{L} - \mathcal{L^*}} .
\end{equation}
Notice that the uniform sampling is used for the training set $\mathcal{S}_\text{selected}$ at the beginning of training as a balancing cold-start strategy.
The corresponding pseudo code can be found in Line \ref{lst:3}.


\section{Experiments}

\begin{table*}[!t]
\centering
{
\centering
\begin{tabular}{l|cc|cc}
\toprule
\multirow{2}{*}{\textbf{Method}} & \multicolumn{2}{c|}{\textbf{M2O}} & \multicolumn{2}{c}{\textbf{O2M}} \\
& \textbf{Related} & \textbf{Diverse} & \textbf{Related} & \textbf{Diverse} \\
\midrule
Bitext Models & 20.37 & 22.38 & 15.73 & 17.83 \\
Uniform Sampling $(\tau = \infty)$ & 22.63 & 24.81 & 15.54 & 16.86 \\
Temperature-Based Sampling $(\tau = 5)$ & 24.00 & 26.01 & 16.61 & 17.94 \\
Proportional Sampling $(\tau = 1)$ & 24.88 & 26.68 & 15.49 & 16.79 \\ 
\midrule
MultiDDS \cite{wang-etal-2020-balancing} & 25.26 & 26.65 & 17.17 & 18.40 \\ 
MultiDDS-S \cite{wang-etal-2020-balancing} & 25.52 & 27.00 & 17.32 & 18.24 \\
\midrule
$\text{CCL-M}_\text{max}$ (Ours) & 26.59** & 28.29** & \textbf{18.89}** & \textbf{19.53}** \\ 
$\text{CCL-M}_\text{avg}$ (Ours) & \textbf{26.73}** & \textbf{28.34}** & 18.74** & \textbf{19.53}** \\
\bottomrule
\end{tabular}
}
\caption{Average BLEU scores (\%) on test sets of the baselines and our methods.
$\text{CCL-M}_\text{max}$ is the CCL-M algorithm using \emph{maximal HRLs-evaluated Competence}, $\text{CCL-M}_\text{max}$ is the CCL-M algorithm using \emph{weighted average HRLs-evaluated Competence}.
Bold indicates the highest value.
"$**$" indicates significantly \citep{koehn-2004-statistical} better than MultiDDS-S with t-test $p < 0.01$.
}
\label{tab:results}
\end{table*}

\subsection{Dataset Setup}

Following \citet{wang-etal-2020-balancing}, we use the 58-languages-to-English TED talks parallel data \cite{qi-etal-2018-pre} to conduct experiments.
Two sets of language pairs with different levels of language diversity are selected: \emph{related} (language pairs with high similarity) and \emph{diverse} (language pairs with low similarity).
Both of them consist of 4 high resource languages (HRLs) and 4 low resource languages (LRLs).

For the \emph{related} language set, we select 4 HRLs (Turkish: "tur", Russian: "rus", Portuguese: "por", Czech, "ces") and its related LRLs (Azerbaijani: "aze", Belarusian: "bel", Glacian: "glg", Slovak: "slk"). For the \emph{diverse} language set, we select 4 HRLs (Greek: "ell", Bulgarian: "bul", French: "fra", Korean: "kor") and 4 LRLs (Bosnian: "bos", Marathi: "mar", Hindi: "hin", Macedonian: "mkd") as \citep{wang-etal-2020-balancing}.
Please refer to Appendix for a more detailed description.

We test two kinds of multilingual machine translation scenarios for each set: 1) \emph{many-to-one} (M2O): translating 8 languages to English; 2) \emph{one-to-many} (O2M): translating English to 8 languages.
The data is preprocessed by SentencePiece\footnote{\url{https://github.com/google/sentencepiece}} \citep{kudo-richardson-2018-sentencepiece} with a vocabulary size of 8k for each language.
Moreover, we add a target language tag before the source and target sentences in O2M as \citep{johnson-etal-2017-googles}.

\subsection{Implementation Details} \label{model}

\paragraph{Baseline.} We select three static heuristic strategies: uniform sampling, proportional sampling, and temperature-based sampling ($\tau = 5$), and the bitext models for the baseline.
In addition, we compare our approach with the previous state-of-the-art sampling method, MultiDDS-S \citep{wang-etal-2020-balancing}. All baseline methods use the same model and the same set of hyper-parameters as our approach.

\paragraph{Model.} We validate our approach upon the multilingual Transformer \citep{vaswani2017attention} implemented by fairseq\footnote{\url{https://github.com/pytorch/fairseq}} \citep{ott-etal-2019-fairseq}.
The number of layers is 6 and the number of attention heads is 4, with the embedding dimension $d_{\text{model}}$ of 512 and the feed-forward dimension $d_{\text{ff}}$ of 1024 as \citep{wang-etal-2020-balancing}.
For training stability, we adopt Pre-LN \citep{xiong2020layer} for the layer-norm \citep{ba2016layer} module.
For M2O tasks, we use a shared encoder with a vocabulary of 64k.
Similarly, for O2M tasks, we use a shared decoder with a vocabulary of 64k.

\paragraph{Training Setup.} We use the Adam optimizer \citep{kingma2014adam} with $\beta_1 = \text{0.9}$, $\beta_2 = \text{0.98}$ to optimize the model.
Further, the same learning rate schedule as \citet{vaswani2017attention} is used, i.e., linearly increase the learning rate for 4000 steps to 2e-4 and decay proportionally to the inverse square root of the step number.
We accumulate the batch size to 9,600 and adopt half-precision training implemented by apex\footnote{\url{https://github.com/NVIDIA/apex}} for faster convergence \citep{ott-etal-2018-scaling}.
For regularization, we also use a dropout \citep{srivastava2014dropout} $p = \text{0.3}$ and a label smoothing \citep{szegedy2016rethinking} $\epsilon_{ls} = \text{0.1}$.
As for our approach, we sample 256 candidates from each languages' development corpora every 100 steps to calculate the \emph{Self-evaluated Competence} $c$ for each language and \emph{HRLs-evaluated Competence} $\hat{c}$ for each LRL.

\paragraph{Evaluation.} \label{sec:eval}
In practice, we perform a grid search for the best threshold $t$ in \{0.5, 0.6, 0.7, 0.8, 0.9, 1.0\}, and select the checkpoints with the lowest weighted loss\footnote{This loss is calculated by averaging the loss of each samples in development corpora of all languages, which is equivalent to taking the proportional weighted average of the loss for each language.} on the development sets to conduct the evaluation.
The corresponding early stopping patience is set to 10.
For target sentence generation, we set the beam size to 5 and a length penalty of 1.0.
Following \citet{wang-etal-2020-balancing}, we use the SacreBLEU \citep{post-2018-call} to evaluate the model performance.
In the end, we compare our result with MultiDDS-S using paired bootstrap resampling \citep{koehn-2004-statistical} for significant test.

\subsection{Results}

\begin{table}[t]
\centering
\resizebox{\columnwidth}{!}{
\centering
\begin{tabular}{l|cc|cc}
\toprule
\multirow{2}{*}{\textbf{Method}} & \multicolumn{2}{c|}{\textbf{Related M2O}} & \multicolumn{2}{c}{\textbf{Diverse M2O}} \\
& \textbf{LRLs} & \textbf{HRLs} & \textbf{LRLs} & \textbf{HRLs} \\
\midrule
Bi. & 10.45 & \textbf{30.29} & 11.18 & \textbf{33.58} \\
MultiDDS-S & 22.51 & 28.54 & 22.72 & 31.29 \\
\midrule
$\text{CCL-M}_\text{max}$ & 23.14* & 30.04** & 23.31* & 33.26** \\ 
$\text{CCL-M}_\text{avg}$ & \textbf{23.30}* & 30.15** & \textbf{23.55}* & 33.13** \\
\bottomrule
\end{tabular}
}
\caption{Average BLEU scores (\%) on test sets of the HRLs and the LRLs for the best baselines and our methods in M2O tasks. Bitext models (``Bi." for short) and MultiDDS-S are selected from the baselines since ``Bi." performs better on the HRLs and MultiDDS-S performs better on the LRLs. Bold indicates the highest value.
"$*$" and "$**$" indicates significantly better than MultiDDS-S with t-test $p < 0.05$ and $p < 0.01$, respectively.
}
\label{tab:m2o hrl and lrl}
\end{table}

\paragraph{Main Results.} \label{main}
The main results are listed in Table \ref{tab:results}.
As we can see, both methods significantly outperform the baselines and MultiDDS with averaged BLEU scores of over +1.07 and +1.13, respectively, indicating the superiority of our approach.
Additionally, the $\text{CCL-M}_\text{avg}$ is slightly better than the $\text{CCL-M}_\text{max}$ in more cases.
This is because the $\text{CCL-M}_\text{avg}$ can get more information provided by the HRLs, and can more accurately estimate when to add an LRL into the training.
Moreover, we find that O2M tasks are much more complicated than M2O tasks since decoders shared by multiple languages might generate tokens in wrong languages.
Consequently, the BLEU scores of O2M tasks are more inferior than M2O tasks by a large margin.

\paragraph{Results on HRLs and LRLs in M2O.}

We further study the performance of our approach on LRLs and the HRLs in M2O tasks and list the results in Table \ref{tab:m2o hrl and lrl}.
As widely known, the bitext model performs poorly on LRLs while performs well on HRLs.
Also, we find our method performs much better than MultiDDS-S, both on LRLs and HRLs.
Although our method does not strictly match the performance of the bitext model on HRLs, the gap between them is much smaller than that of MultiDDS-S and bitext models.
All of the above proves the importance of balancing learning competencies of different languages.

\paragraph{Results on HRLs and LRLs in O2M.}

\begin{table}[t]
\centering
\resizebox{\columnwidth}{!}{
\centering
\begin{tabular}{l|cc|cc}
\toprule
\multirow{2}{*}{\textbf{Method}} & \multicolumn{2}{c|}{\textbf{Related O2M}} & \multicolumn{2}{c}{\textbf{Diverse O2M}} \\
& \textbf{LRLs} & \textbf{HRLs} & \textbf{LRLs} & \textbf{HRLs} \\
\midrule
Bi. & 8.25 & \textbf{23.22} & 7.82 & \textbf{27.83} \\
MultiDDS-S & 15.31 & 19.34 & 13.98 & 22.52 \\
\midrule
$\text{CCL-M}_\text{max}$ & \textbf{16.54}** & 21.24** & \textbf{14.36}* & 24.71** \\ 
$\text{CCL-M}_\text{avg}$ & 16.33** & 21.14** & 13.82 & 25.42** \\
\bottomrule
\end{tabular}
}
\caption{Average BLEU scores (\%) on test sets of the HRLs and the LRLs for the best baselines and our methods in O2M tasks.
Bitext models (``Bi." for short) and MultiDDS-S are selected from the baselines since ``Bi." performs better on the HRLs and MultiDDS-S performs better on the LRLs. Bold indicates the highest value.
"$*$" and "$**$" indicates significantly better than MultiDDS-S with t-test $p < 0.05$ and $p < 0.01$, respectively.
}
\label{tab:o2m hrl and lrl}
\end{table}

\begin{figure*}[!t]
\centering
\includegraphics[width=0.9\textwidth]{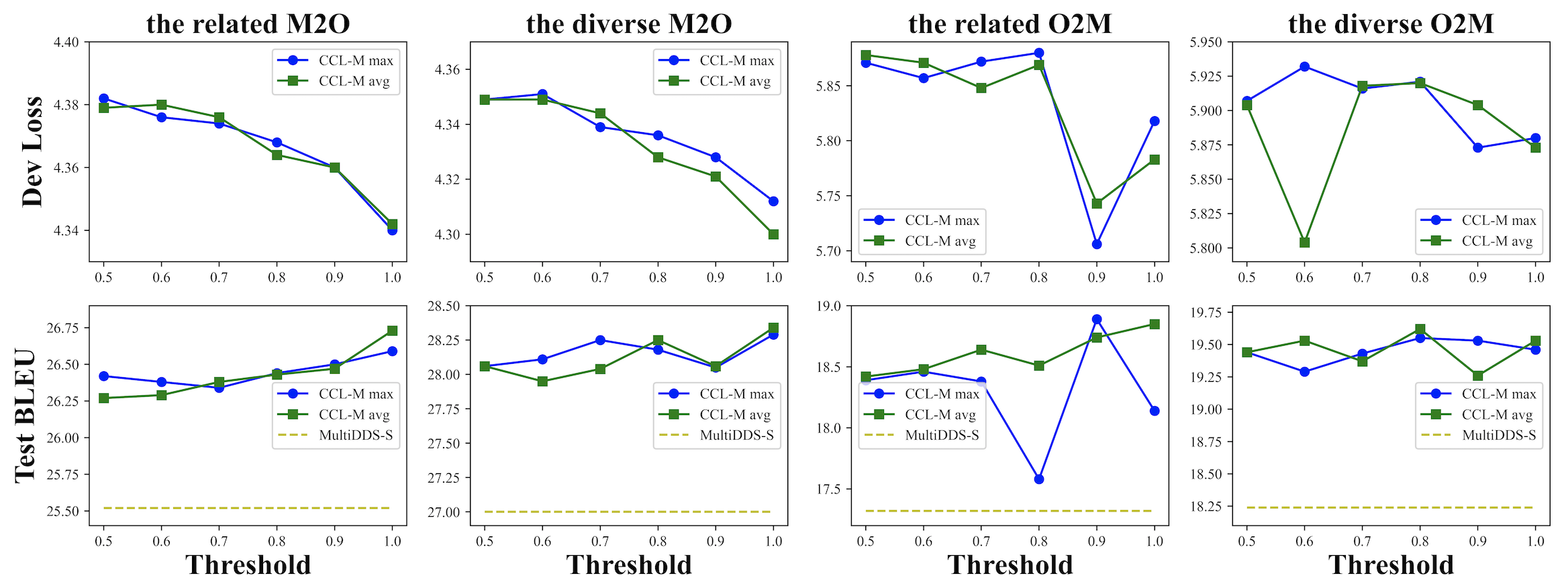}
\caption{Weighted losses on development sets and average BLEU scores (\%) on test sets for different thresholds (the abscissa) in four scenarios. The blue line represents $\text{CCL-M}_\text{max}$, the green line represents $\text{CCL-M}_\text{avg}$. 
The yellow dotted line represents MultiDDS-S \citep{wang-etal-2020-balancing}.}
\label{grid}
\end{figure*}

As shown in Table \ref{tab:o2m hrl and lrl}, our approach also performs well on the more difficult scenario, i.e., the O2M.
Apparently, our approach almost doubles the performance of the LRLs from bitext models.
Consistently, there is a roughly -2 and -3 BLEU decay for the HRLs in the \emph{related} and \emph{diverse} language sets.
Compared to MultiDDS-S, both our approach in the LRLs and the HRLs are significantly better.
This again proves the importance of balancing the competencies of different languages.
Additionally, the performance on HRLs in O2M task has a larger drop from the bitext model than that in M2O task.
This is because the decoder shares a 64k vocabulary for all languages in O2M tasks, but each language has only 8k vocabulary.
Thus, it is easier for the model to output misleading tokens that do not belong to the target language during inference.

\section{Analysis}

\subsection{Effects of Different Threshold $t$} 
\label{threshold}

We firstly conduct a grid search for the best \emph{HRLs-evaluated Competence} threshold $t$.
As we can see from Figure \ref{grid}, the more HRLs are trained (the larger the threshold $t$ is), the better the model's performance is in M2O tasks.
This phenomenon again suggests that M2O tasks are easier than O2M tasks.
The curriculum learning framework performs better in the \emph{related} set than that in the \emph{diverse} set in M2O tasks, because languages in the \emph{related} set are more similar.
Still, our method is better than MultiDDS-S, as shown in Figure \ref{grid}.
This again demonstrates the positive effect of our curriculum learning framework.

Experimental results also reveal that the optimal threshold $t$ for O2M tasks may not be 1 because more training on HRLs would not produce optimal overall performance.
Furthermore, the optimal threshold for the \emph{diverse} language set is lower than that for the \emph{related} language set as the task in the \emph{diverse} language set is more complicated.

\subsection{Effects of Different Sampling Methods}

\begin{table}[t]
\resizebox{\columnwidth}{!} {
\centering
\begin{tabular}{l|cc|cc}
\toprule
\multirow{2}{*}{\textbf{Method}} & \multicolumn{2}{c|}{\textbf{M2O}} & \multicolumn{2}{c}{\textbf{O2M}} \\
& \textbf{Related} & \textbf{Diverse} & \textbf{Related} & \textbf{Diverse} \\
\midrule
$\text{CCL-M}_\text{avg}$ & 26.73 & 28.34 & \textbf{18.74} & \textbf{19.53} \\
\ \ \ $+$ Uni. & 24.59 & 27.13 & 18.29 & 18.21 \\
\ \ \ $+$ Temp. & 25.28 & 27.50 & 18.65 & 19.28 \\
\ \ \ $+$ Prop. & \textbf{27.21} & \textbf{28.72} & 18.20 & 18.80 \\ 
\bottomrule
\end{tabular}
}
\caption{Average BLEU scores (\%) on test sets by the $\text{CCL-M}_\text{avg}$ algorithm using our dynamic sampling method and three static sampling methods. "Uni." refers to the uniform sampling, "Temp." refers to the temperature-based sampling ($\tau = 5$), and "Prop." refers to the proportional sampling. Bold indicates the highest value.
}
\label{tab:sample}
\end{table}

We also analyze the effects of different sampling methods.
Substituting our competence-aware dynamic sampling method in the $\text{CCL-M}_\text{avg}$ with three static sampling methods, we get the results in Table \ref{tab:sample}.
Consistently, our method performs best among the sampling methods in O2M tasks, which shows the superiority of sampling by language-specific competence.

Surprisingly, we find that proportional sampling surpasses our proposed dynamic method in M2O tasks.
This also indicates that more training on HRLs has a positive effect in M2O tasks, since proportional sampling would train more on the HRLs than the dynamic sampling we proposed.
In addition, all three static sampling methods outperform their respective baselines in Table \ref{tab:results}. Some of them are even better than the previous state-of-the-art sampling method, i.e., MultiDDS-S.
This shows that our curriculum learning approach has a strong generability.






\section{Related Work}

Curriculum learning was first proposed by \citet{bengio2009curriculum} with the idea of learning samples from easy to hard to get a better optimized model.
As a general method for model improvement, curriculum learning has been widely used in a variety of machine learning fields \citep{gong2016multi, kocmi-bojar-2017-curriculum, hacohen2019power, platanios-etal-2019-competence, narvekar2020curriculum}

There are also some previous curriculum learning researches for machine translation.
For example, \citep{kocmi-bojar-2017-curriculum} divide the training corpus into smaller buckets using some features such as sentence length or word frequency and then train the buckets from easy to hard according to the predefined difficulty.
\citet{platanios-etal-2019-competence} propose competence-based curriculum learning for machine translation, which treats the model competence as a variable in training and samples the training corpus in line with the competence.
In detail, they believes that competence is positively related to the training steps, and uses linear or square root functions for experiments.
We bring the concept of competence and redefine it in this paper with a multilingual context.
Further, we define \emph{Self-evaluated Competence} and \emph{HRLs-evaluated Competence} as the competence of each language pair to capture the model's multilingual competence more accurately.

\section{Conclusion}

In this paper, we focus on balancing the learning competencies of different languages in multilingual machine translation and propose a competence-based curriculum learning framework for this task.
The experimental results show that our approach brings significant improvements over baselines and the previous state-of-the-art balancing sampling method, MultiDDS-S.
Furthermore, the ablation study on sampling methods verifies the great generalibility of our curriculum learning framework.

\section*{Acknowledgements}
We would like to thank anonymous reviewers for their suggestions and comments. This work was supported by the National Key Research and Development Program of China (No. 2020YFB2103402).

\bibliography{anthology,custom}
\bibliographystyle{acl_natbib}

\appendix

\section{Dataset}
\label{sec:statistics}

\subsection{Dataset Statistics}

As we can see in Table \ref{tab:related statistics} and Table \ref{tab:diverse statistics}, there are 4 low resource languages (LRLs) and 4 high resource languages (HRLs) in both language sets.

\begin{table}[h]
\centering
\begin{tabular}{c|ccc}
\toprule
\textbf{Language} & \textbf{Train} & \textbf{Dev} & \textbf{Test} \\
\midrule
aze & 5.94k & 671  & 903  \\
bel & 4.51k & 248  & 664  \\
glg & 10.0k & 682  & 1007 \\ 
slk & 61.5k & 2271 & 2445 \\ 
tur & 182k  & 4045 & 5029 \\
rus & 208k  & 4814 & 5483 \\ 
por & 185k  & 4035 & 4855 \\
ces & 103k  & 3462 & 3831 \\
\bottomrule
\end{tabular}
\caption{Dataset statistics of the \emph{related} language set.}
\label{tab:related statistics}
\end{table}

\begin{table}[h]
\centering
\begin{tabular}{c|ccc}
\toprule
\textbf{Language} & \textbf{Train} & \textbf{Dev} & \textbf{Test} \\
\midrule
bos & 5.64k & 474  & 463  \\
mar & 9.84k & 767  & 1090 \\
hin & 18.7k & 854  & 1243 \\ 
mkd & 25.3k & 640  & 438  \\ 
ell & 134k  & 3344 & 4433 \\
bul & 174k  & 4082 & 5060 \\ 
fra & 192k  & 4320 & 4866 \\
kor & 205k  & 4441 & 5637 \\
\bottomrule
\end{tabular}
\caption{Dataset statistics of the \emph{diverse} language set.}
\label{tab:diverse statistics}
\end{table}

\subsection{Development Losses}

We use the same model and hyper-parameters as we used in subsection \ref{model} and get the results in Table \ref{tab:related losses} and Table \ref{tab:diverse losses}.
We then use them to calculate the \emph{Self-evaluated Competence}.
Obviously, the losses of HRLs are lower than the losses of LRLs.
At the same time, we find that the \emph{xxx-eng} tasks is easier than the \emph{eng-xxx} tasks.
Because in \emph{eng-xxx} tasks, the decoder shares a 64k vocabulary and would output misleading tokens.

\begin{table}[!t]
\centering
\begin{tabular}{c|cc}
\toprule
\textbf{Language} & \textbf{xxx-eng} & \textbf{eng-xxx} \\
\midrule
aze & 7.87  & 9.703 \\
bel & 7.843 & 9.051 \\
glg & 6.891 & 7.688 \\ 
slk & 5.205 & 5.84  \\ 
tur & 4.344 & 5.225 \\
rus & 4.577 & 5.011 \\ 
por & 3.687 & 4.067 \\
ces & 4.495 & 5.083 \\
\bottomrule
\end{tabular}
\caption{Losses on development sets of bitext models in the \emph{related} language set.}
\label{tab:related losses}
\end{table}

\begin{table}[!t]
\centering
\begin{tabular}{c|cc}
\toprule
\textbf{Language} & \textbf{xxx-eng} & \textbf{eng-xxx} \\
\midrule
bos & 7.499 & 8.687 \\
mar & 7.472 & 9.184 \\
hin & 6.956 & 7.961 \\ 
mkd & 5.581 & 6.221 \\ 
ell & 4.164 & 4.522 \\
bul & 4.004 & 4.278 \\ 
fra & 3.883 & 3.968 \\
kor & 4.725 & 5.843 \\
\bottomrule
\end{tabular}
\caption{Losses on development sets of bitext models in the \emph{diverse} language set.}
\label{tab:diverse losses}
\end{table}

\begin{table}[!t]
\centering
\begin{tabular}{c|cccc}
\toprule
\textbf{Language} & aze & bel & glg & slk \\
\midrule
tur & \textbf{0.50} & 0.12 & 0.24 & 0.30 \\
rus & 0.09 & \textbf{0.34} & 0.07 & 0.08 \\ 
por & 0.22 & 0.12 & \textbf{0.59} & 0.26 \\
ces & 0.24 & 0.11 & 0.27 & \textbf{0.68} \\
\bottomrule
\end{tabular}
\caption{Language similarity of the \emph{related} language set. Bold indicates significant similarity.}
\label{tab:related similarity}
\end{table}

\begin{table}[!t]
\centering
\begin{tabular}{c|cccc}
\toprule
\textbf{Language} & bos & mar & hin & mkd \\
\midrule
ell & 0.09 & 0.09 & 0.09 & 0.07 \\
bul & 0.12 & 0.11 & 0.11 & \textbf{0.60} \\ 
fra & 0.18 & 0.08 & 0.09 & 0.07 \\
kor & 0.10 & 0.10 & 0.09 & 0.07 \\
\bottomrule
\end{tabular}
\caption{Language similarity of the \emph{diverse} language set. Bold indicates significant similarity.}
\label{tab:diverse similarity}
\end{table}

\subsection{Language Similarity}

Using Equation \ref{eq:sim}, we obtain the language similarities shown in Table \ref{tab:related similarity} and Table \ref{tab:diverse similarity}.
As we can see, each HRL has a high-similarity LRL corresponding to it in the \emph{related} language set.
Meanwhile, languages are generally not similar in \emph{diverse} language set, only a pair of HRL and LRL ("bul" and "mkd") have high similarity.

\section{Individual BLEU Scores}

Here we also list the individual BLEU scores as the supplement to Table \ref{tab:results}.

\begin{table*}[h!]
\centering
\scalebox{0.85}{
\begin{tabular}{l|c|cccccccc}
\toprule
\textbf{Method} & \textbf{SacreBLEU} & \textbf{aze} & \textbf{bel} & \textbf{glg} & \textbf{slk} & \textbf{tur} & \textbf{rus} & \textbf{por} & \textbf{ces} \\
\midrule
Bitext Models & 22.38 & 7.07 & 3.77 & 10.79 & 23.09 & 37.33 & \textbf{38.26} & \textbf{39.75} & \textbf{18.96} \\
Uniform Sampling $(\tau = \infty)$ & 24.81 & 21.52 & 9.48 & 19.99 & 30.46 & 33.22 & 33.70 & 35.15 & 15.03 \\
Temperature-Based Sampling $(\tau = 5)$ & 26.01 & 23.47 & 10.19 & 21.26 & 31.13 & 34.69 & 34.94 & 36.44 & 16.00 \\
Proportional Sampling $(\tau = 1)$ & 26.68 & 23.43 & 10.10 & 22.01 & 31.06 & 35.62 & 36.41 & 37.91 & 16.91 \\ 
\midrule
MultiDDS \cite{wang-etal-2020-balancing} & 26.65 & 25.00 & 10.79 & 22.40 & 31.62 & 34.80 & 35.22 & 37.02 & 16.36 \\ 
MultiDDS-S \cite{wang-etal-2020-balancing} & 27.00 & 25.34 & 10.57 & 22.93 & 32.05 & 35.27 & 35.77 & 37.30 & 16.81 \\
\midrule
$\text{CCL-M}_\text{max}$ (Ours) & 28.29 & 25.20 & 11.50 & 23.23 & 33.31 & \textbf{37.55} & 38.03 & 39.26 & 18.20 \\ 
$\text{CCL-M}_\text{avg}$ (Ours) & \textbf{28.34} & \textbf{25.61} & \textbf{11.60} & \textbf{23.52} & \textbf{33.48} & 37.26 & 38.10 & 39.19 & 17.98 \\
\bottomrule
\end{tabular}
}
\caption{Individual BLEU scores (\%) on test sets of the baselines and our methods in M2O diverse language sets. Bold indicates the highest value.}
\label{tab:m2o diverse individual}
\end{table*}

\begin{table*}[h!]
\centering
\scalebox{0.85}{
\begin{tabular}{l|c|cccccccc}
\toprule
\textbf{Method} & \textbf{SacreBLEU} & \textbf{aze} & \textbf{bel} & \textbf{glg} & \textbf{slk} & \textbf{tur} & \textbf{rus} & \textbf{por} & \textbf{ces} \\
\midrule
Bitext Models & 20.37 & 2.59 & 2.69 & 11.62 & 24.88 & \textbf{26.34} & \textbf{24.12} & \textbf{44.53} & 26.18 \\
Uniform Sampling $(\tau = \infty)$ & 22.63 & 8.81 & 14.80 & 25.22 & 27.32 & 20.16 & 20.95 & 38.69 & 25.11 \\
Temperature-Based Sampling $(\tau = 5)$ & 24.00 & 10.42 & 15.85 & 27.63 & 28.38 & 21.53 & 21.82 & 40.18 & 26.26 \\
Proportional Sampling $(\tau = 1)$ & 24.88 & 11.20 & 17.17 & 27.51 & 28.85 & 23.09 & 22.89 & 41.60 & 26.80 \\ 
\midrule
MultiDDS \cite{wang-etal-2020-balancing} & 25.26 & 12.20 & 18.60 & 28.83 & 29.21 & 22.24 & 22.50 & 41.40 & 27.22 \\ 
MultiDDS-S \cite{wang-etal-2020-balancing} & 25.52 & 12.20 & 19.11 & 29.37 & 29.35 & 22.81 & 22.78 & 41.55 & 27.03 \\
\midrule
$\text{CCL-M}_\text{max}$ (Ours) & 26.59 & \textbf{12.61} & 19.43 & 29.96 & 30.55 & 24.63 & 23.93 & 43.05 & 28.55 \\ 
$\text{CCL-M}_\text{avg}$ (Ours) & \textbf{26.73} & 12.59 & \textbf{19.54} & \textbf{30.20} & \textbf{30.86} & 24.78 & 24.09 & 43.13 & \textbf{28.61} \\
\bottomrule
\end{tabular}
}
\caption{Individual BLEU scores (\%) on test sets of the baselines and our methods in M2O related language sets. Bold indicates the highest value.}
\label{tab:m2o related individual}
\end{table*}

\begin{table*}[h!]
\centering
\scalebox{0.85}{
\begin{tabular}{l|c|cccccccc}
\toprule
\textbf{Method} & \textbf{SacreBLEU} & \textbf{bos} & \textbf{mar} & \textbf{hin} & \textbf{mkd} & \textbf{ell} & \textbf{bul} & \textbf{fra} & \textbf{kor} \\
\midrule
Bitext Models & 15.73 & 2.22 & 2.54 & 9.82 & 18.40 & \textbf{15.02} & \textbf{19.57} & \textbf{39.42} & 18.86 \\
Uniform Sampling $(\tau = \infty)$ & 15.54 & 5.76 & 10.51 & 21.08 & 17.83 & 9.94 & 13.59 & 30.33 & 15.35 \\
Temperature-Based Sampling $(\tau = 5)$ & 16.61 & 6.66 & 11.29 & 21.81 & 18.60 & 11.27 & 14.92 & 32.10 & 16.26 \\
Proportional Sampling $(\tau = 1)$ & 15.49 & 4.42 & 5.99 & 14.92 & 17.37 & 12.86 & 16.98 & 34.90 & 16.53 \\ 
\midrule
MultiDDS \cite{wang-etal-2020-balancing} & 17.17 & 6.24 & 11.75 & 21.46 & 20.67 & 11.51 & 15.42 & 33.41 & 16.94 \\ 
MultiDDS-S \cite{wang-etal-2020-balancing} & 17.32 & 6.59 & 12.39 & 21.65 & 20.61 & 11.58 & 15.26 & 33.52 & 16.98 \\
\midrule
$\text{CCL-M}_\text{max}$ (Ours) & \textbf{18.89} & \textbf{7.59} & 13.01 & 23.83 & \textbf{21.71} & 13.38 & 16.85 & 35.43 & \textbf{19.31} \\ 
$\text{CCL-M}_\text{avg}$ (Ours) & 18.85 & 7.38 & \textbf{13.17} & \textbf{23.89} & 21.67 & 13.37 & 16.92 & 35.34 & 19.04 \\
\bottomrule
\end{tabular}
}
\caption{Individual BLEU scores (\%) on test sets of the baselines and our methods in O2M related language sets. Bold indicates the highest value.}
\label{tab:o2m related individual}
\end{table*}

\begin{table*}[h!]
\centering
\scalebox{0.85}{
\begin{tabular}{l|c|cccccccc}
\toprule
\textbf{Method} & \textbf{SacreBLEU} & \textbf{bos} & \textbf{mar} & \textbf{hin} & \textbf{mkd} & \textbf{ell} & \textbf{bul} & \textbf{fra} & \textbf{kor} \\
\midrule
Bitext Models & 17.83 & 5.00 & 2.68 & 8.17 & 15.44 & \textbf{31.35} & \textbf{33.88} & \textbf{38.02} & \textbf{8.06} \\
Uniform Sampling $(\tau = \infty)$ & 16.86 & 14.12 & 4.69 & 14.52 & 20.10 & 22.87 & 25.02 & 27.64 & 5.95 \\
Temperature-Based Sampling $(\tau = 5)$ & 17.94 & 14.73 & 4.93 & 15.49 & 20.59 & 24.82 & 26.60 & 29.74 & 6.62 \\
Proportional Sampling $(\tau = 1)$ & 16.79 & 6.93 & 3.69 & 10.70 & 15.77 & 26.69 & 29.59 & 33.51 & 7.49 \\ 
\midrule
MultiDDS \cite{wang-etal-2020-balancing} & 18.40 & \textbf{14.91} & \textbf{4.83} & 14.96 & 22.25 & 24.80 & 27.99 & 30.77 & 6.76 \\ 
MultiDDS-S \cite{wang-etal-2020-balancing} & 18.24 & 14.02 & 4.76 & 15.68 & 21.44 & 25.69 & 27.78 & 29.60 & 7.01 \\
\midrule
$\text{CCL-M}_\text{max}$ (Ours) & \textbf{19.53} & 14.87 & 4.81 & \textbf{15.33} & \textbf{22.43} & 28.10 & 29.97 & 33.31 & 7.44 \\ 
$\text{CCL-M}_\text{avg}$ (Ours) & \textbf{19.53} & 14.87 & 4.81 & \textbf{15.33} & \textbf{22.43} & 28.10 & 29.97 & 33.31 & 7.44 \\
\bottomrule
\end{tabular}
}
\caption{Individual BLEU scores (\%) on test sets of the baselines and our methods in O2M diverse language sets. Bold indicates the highest value. The scores of $\text{CCL-M}_\text{max}$ and $\text{CCL-M}_\text{avg}$ are the same because they add different LRLs to the training set at the same time (the training set $\mathcal{S}_\text{selected}$ is adjusted every 100 steps), even with different thresholds.}
\label{tab:o2m diverse individual}
\end{table*}

\end{document}